# Person Re-identification in Appearance Impaired Scenarios

Mengran Gou*, Xikang Zhang**, Angels Rates-Borras**
Sadjad Asghari-Esfeden**, Mario Sznaier*, Octavia Camps*
Robust Systems Lab, Northeastern University
Boston, USA
*{mengran, msznaier, camps}@coe.neu.edu
**{zhang.xik, ratesborras.a, asghariesfeden.s}@husky.neu.edu

## Abstract

*Person re-identification is critical in surveillance applications. Current approaches rely on appearance based features extracted from a single or multiple shots of the target and candidate matches. These approaches are at a disadvantage when trying to distinguish between candidates dressed in similar colors or when targets change their clothing. In this paper we propose a dynamics-based feature to overcome this limitation. The main idea is to capture soft biometrics from gait and motion patterns by gathering dense short trajectories (tracklets) which are Fisher vector encoded. To illustrate the merits of the proposed features we introduce three new "apperance-impaired" datasets. Our experiments on the original and the appearance impaired datasets demonstrate the benefits of incorporating dynamics-based information with appearance-based information to re-identification algorithms.*

## 1. Introduction

Visual surveillance using cameras with little or no overlapping field of view requires addressing the problem of human re-identification (re-id) [6, 13, 22, 36, 5, 3, 14]. Re-id is a very challenging problem due to significant variations in appearance, caused by changes in viewpoint, illumination and pose. In some scenarios, targets may reappear a few days later wearing different clothes. One approach to address this challenge is to learn a mapping function such that the distances between features from the same person are relatively small, while the distances between features from different persons are relatively large [35, 16, 29, 39, 18]. Other approaches focus on designing robust and invariant descriptors to better represent the subjects across views [4, 37, 25, 24, 2]. By intelligently fusing different types of appearance features, [38, 28, 9] achieve state-of-the-art performance on several datasets.

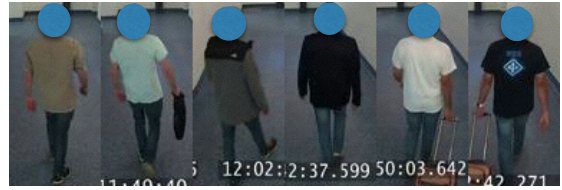

Figure 1. Two examples of images of the same person but wearing different clothing. Each triplet of images shows the same person.

Currently, most re-id methods rely on appearance based features such as color and texture statistics which are extracted either from a single image or from a small set of images of the targets. These types of features are at disadvantage when matching two views of the same person but wearing different clothing (for example images before and after removing a coat as shown in Figure 1) or when trying to distinguish people wearing similar clothing (for example pedestrians wearing black business suits as shown in Figure 2).

Most of the re-id literature work with still images. However, in real surveillance applications, the vision system has the ability to track the individuals for a while [19, 7], enabling the possibility of incorporating temporal/dynamic information to aid re-identification in these difficult scenarios. Yet, very few approaches take advantage of this capability. For example, in [18], although re-id starts from frame sequences, the procedure iteratively applies Fisher analysis and hierarchical clustering to obtain a set of most discriminative frames and discards the temporal information. Notably, Liu *et al*. [23] proposed to use Gait Energy Image (GEI) [12] features to re-identify persons with different appearances. However, obtaining good gait silhouettes is extremely hard in crowded scenes such as airport terminals. Wang *et al*. [33] used spatial-temporal HOG3D features [15] and performed re-id by selecting and ranking discriminating video fragments, simultaneously.

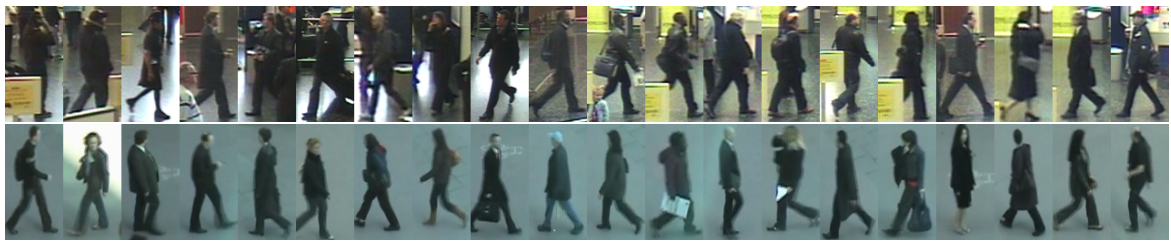

Figure 2. Examples of persons wearing black suits. The first row was collected from the iLIDSVID dataset and the second row was collected from the PRID dataset.

In this paper, inspired by recent results in cross-view activity recognition [17, 11] and the success of Fisher vector encoding approaches [31], we propose to capture soft biometrics such as gait and motion patterns by Fisher vector encoding temporal pyramids of dense short trajectories.

To evaluate the benefits of the proposed features, we compiled three new challenging "appearance impaired" re-id datasets. Two of these are subsets of the iLIDSVID and PRID2011 datasets, and are entirely comprised of videos with people wearing black clothes. The third set, collected by us, was captured using surveillance cameras from a train station. This set has video sequences where the same people appear wearing different clothing and accessories. Our experiments, using the full standard re-id datasets as well as the ones introduced here, show that combining the proposed feature with existing appearance-based features improves re-id performance overall, but more so in appearance-impaired situations as the ones described above.

Thus, the main contributions of this paper are:

- A novel dynamics-based feature Fisher vector encoding DynFV for re-id. The proposed feature captures subtle motion patterns to aid re-id, and in particular in appearance-impaired scenarios.

- Three new challenging "appearance impaired" datasets for re-id performance evaluation.

- A comprehensive evaluation of the effect of choosing different spatio, spatio-temporal, and dynamics-based features on the performance of (unsupervised) re-id methods.

## 2. Related Work

Appearance based features capturing spatial characteristic such as color and texture have been widely used in person re-identification studies [5]. Xiong *et al.* [35] used color histograms and LBP features to evaluated the effect of using different spatial splitting schema to compute them. Covariance matrices have also been used as features by [24, 2]. Zhao *et al.* [36] used SIFT features to introduce a re-id method based on salience parts. Bazzani *et al.* [4] proposed a global appearance model SDALF using three complementary visual features. Although SDALF can be built based on multi-shots, it only uses spatial data.

On the other hand, the use of temporal information in the re-id literature is very limited. Gheissari *et al.* [10] incorporated temporal information to extract salient edge features for person re-id. By assuming that the person has been tracked for several frames, [3] used the trajectory to compensate the viewpoints variance. Wang *et al.* [33] split the video sequence into several candidate fragments based on motion energy and then extracted time-space HOG3D features[15] for each of the fragments. Then, by applying a multi-instance ranking method, they can select and rank the most discriminating fragments simultaneously. After combining them with the spatial features proposed in [14], they achieved the state of the art performance on the iLIDSVID and PRID datasets.

Dense trajectories capturing temporal information and moving patterns [32] have been proved to be powerful features for activity recognition [30, 34]. More recently, Li *et al.* [17], while addressing the problem of cross-view activity recognition, proposed to encode dense trajectories using Hankelet descriptors. There, they showed that Hankelets carry useful viewpoint and initial condition invariant properties. However, until now, neither dense trajectories nor Hankelets have been used for human re-id.

Re-identification performance can also be improved by using better ways to compare or classify the features being used. [39] proposed a relative distance comparison model to maximize the likelihood of distances between true matches being smaller than distance between wrong matches. Li *et al.* [20] proposed a method to learn a local adaptive threshold for the distance from the metric. In [35], Xiong *et al.* reported a comprehensive evaluation on several metric learning algorithms and presented extensions to PCCA [26] using a regularized term, and to LDA [29] using a kernel trick. More recently, [21] applied cross-view quadratic discriminant analysis (XQDA) to learn the metric on a more discriminative subspace.

Alternatively, Fisher vector encoding methods can combine discriminating descriptors with generative models [31]

providing excellent recognition performance. For example, [34] showed that Fisher vectors are one of the best encoding methods for activity recognition while [30] showed that a two-layer Fisher vector incorporating mid-level Fisher vectors representing semantic information achieved even better performance.

## 3. The DynFV Feature

One of the main objectives of this paper is to address the problem of re-identification in appearance impaired scenarios such as the ones illustrated in Figures 1 and 2. In such cases, gait and idiosyncratic motion patterns offer a natural complementary source of information that is not affected by the lack of discriminating appearance-based features.

However, reliable estimation of motion-based biometrics, such as gait, is very challenging in crowded surveillance videos. In particular, it is very difficult to locate and consistently track the joints of the targets which would be required to model their gait. Because of this, we propose to use instead soft-biometric characteristics provided by *sets of dense, short trajectories (tracklets)*, which have been shown to carry useful invariants [17].

A potential drawback of using dense tracklets is that there are many of them and that they can exhibit large variability. Thus, it is important to have an effective way to aggregate the information they could provide. Towards this goal, we propose to use *pyramids of dense trajectories* with *Fisher vector encoding*, as illustrated in Figure 3 and described in detail below.

### 3.1. Fisher Vector Encoding

The Fisher vector encoding method was first introduced to solve large scale image classification problems [31] and has significantly improved the performance of action recognition systems [30, 34]. As shown in our experiments, using this method allows us to aggregate multiple dynamic (and spatial) features in an effective way. Next, for the sake of completeness, we briefly summarize the main concepts of Fisher vector encoding.

Let $X = \{x_1, x_2, ..., x_N\}$ be a set of feature vectors that can be modeled by a distribution $p(X|\theta)$ with parameters $\theta$. Then, this set can be represented by the gradient vector of the log-likelihood w.r.t the parameters $\theta$. Following the assumptions in [31], $P(X|\theta)$ is modeled using a Gaussian mixture model (GMM) with $\theta = \{\pi_1, \mu_1, \Sigma_1, ..., \pi_k, \mu_k, \Sigma_k\}$, where $k$ is the number of mixture models, $\pi_i$, $\mu_i$, and $\Sigma_i$ are the mixture weight, mean and covariance of Gaussian $i$. Assuming all covariances are diagonal matrices $\Sigma_i = \sigma_i I$, this set of feature vectors can be encoded by equations (1) and (2).

$$F_{\mu,i}^X = \frac{1}{N\sqrt{\pi_i}} \sum_{n=1}^{N} \gamma_n(i) \left( \frac{x_n - \mu_i}{\sigma_i} \right) \quad (1)$$

$$F_{\sigma,i}^X = \frac{1}{N\sqrt{2\pi_i}} \sum_{n=1}^{N} \gamma_n(i) \left\{ \frac{(x_n - \mu_i)^2}{\sigma_i^2} - 1 \right\} \quad (2)$$

where $\gamma_n(k)$ is the soft assignment of $x_n$ to the $k^{th}$ Gaussian model. Then, the Fisher vector of set $X$ will be the concatenation of $F_{\mu,i\in\{1,2,...k\}}^X$ and $F_{\sigma,i\in\{1,2,...k\}}^X$.

### 3.2. Fisher Vector Encoding of Dense Trajectories

Given a short tracklet $(\mathbf{Z}_1, \mathbf{Z}_2, ..., \mathbf{Z}_N)$, its Hankelet [17] is defined as the Hankel matrix:

$$\mathbf{H_Z} = \begin{bmatrix} \mathbf{Z}_1 & \mathbf{Z}_2 & \ldots & \mathbf{Z}_k \\ \mathbf{Z}_2 & \mathbf{Z}_3 & \ldots & \mathbf{Z}_{k+1} \\ \vdots & \vdots & \ldots & \vdots \\ \mathbf{Z}_l & \mathbf{Z}_{l+1} & \ldots & \mathbf{Z}_N \end{bmatrix} \quad (3)$$

The constant off-diagonal structure of the Hankelet (depicted by elements painted with the same color in Figure 4(a)) carries properties that are invariant to viewpoint changes and initial conditions [17].

Here, we propose to capture the underlying dynamic information in the dense trajectories by building a family of Hankelets (with an increasing number of columns $a$) for each trajectory. The rows of these Hankelets are obtained by splitting each trajectory of length $l$ into shorter and shorter tracklets, using an sliding window with full overlap (stride of one) as illustrated in Figure 4, which in turn are Fisher vector encoded (see Figure 3).

Intuitively, the rows of the Hankelets, i.e. short tracklets of increasing length, constitute a set of temporal pyramids capturing different levels of dynamic complexity (since higher order dynamics are represented using longer bigger Hankelets/longer tracklets). More precisely, assume we have a set of pairs of sequences $\{S_a^i, S_b^i\}_{i=1}^{N}$, where $S_a^i$ represents the sequence of person $i$ in camera $a$ and $N$ is the total number of persons. Since different human body parts may have different dynamics, we split the bounding box of the person into $G$ grids and process each grid separately. Then, the temporal pyramid of dynamics-based features is built and encoded using the following steps:

For each grid $g = 1, ..., G$, do:

1. $m$ = Number of tracklets in grid $g^1$ of length $l + 1$ extracted from all sequence pairs in training data

2. Following [17] we work with velocity tracklets rather than position: $V = [\Delta Z_t, \Delta Z_{t+1}, ..., \Delta Z_{t+l}]$ where $\Delta Z_t = [x_{t+1} - x_t, y_{t+1} - y_t]$ and $[x_t, y_t]$ are the coordinates of the tracked point at time $t$.

---
[1]Location of each tracklet is determined by its starting point.

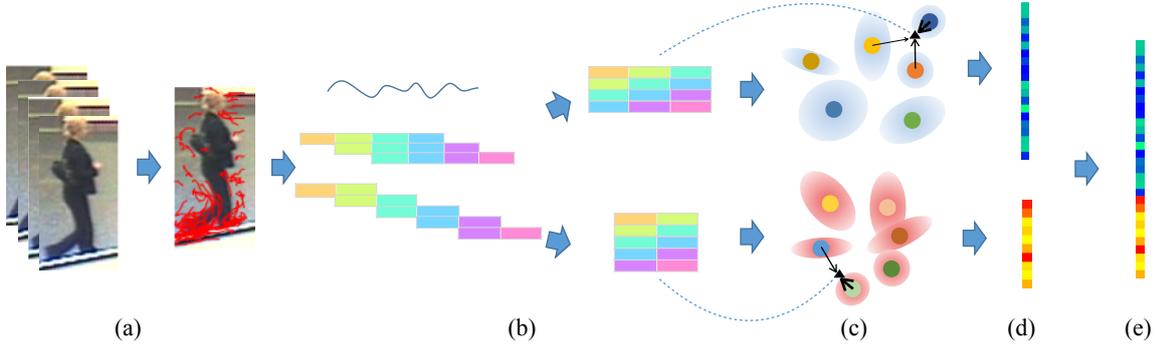

Figure 3. Pipeline of the proposed dynamics-based feature extraction. (a) Dense trajectories are extracted from video sequences and divided into small grids. (b) Temporal pyramids of the original trajectories are built using sliding windows of different sizes. (b-c) A GMM model is learned for each level of the pyramid. (c-d) The trajectories at each level of the pyramid are encoded using Fisher vectors based on the corresponding GMM. (d-e)The Fisher vectors at all scales are pooled to obtain the final feature vector.

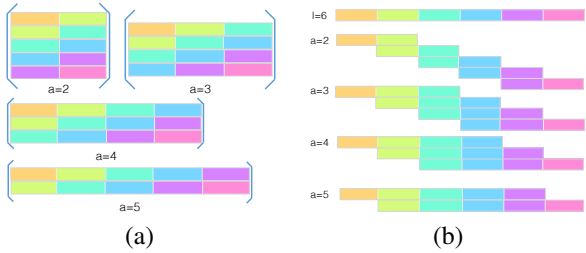

Figure 4. (a) Each trajectory is associated a family of Hankelets, with increasing number of columns, $a$. The rows of these Hankelets are obtained by splitting the trajectory of length $l$ into shorter tracklets of length $a$ using a sliding window with full overlap (stride of 1), as shown in (b).

3. For each pyramid level corresponding to window size $a_i \in \mathcal{A}$ do:

    (a) Generate $m(l - a_i + 1)$ shorter tracklets set $V_{a_i}{}^g$ of length $a_i$ using a sliding window of size $a_i$ and stride of 1.

    (b) Learn a GMM model $P$ for the tracklets set $V_{a_i}{}^g$.

    (c) Encode these tracklets using a Fisher vector $F_P{}^{V_{a_i}^g}$ based on the obtained GMM model.

    (d) Apply power normalization, followed by L2 normalization.

    Pool Fisher vectors by concatenating all $F_P{}^{V_{a_i}^g}$.

## 4. Datasets for Performance Evaluation

The second major objective of this paper is to present a thorough performance evaluation of multi-shot unsupervised re-identification systems as different features are used. We evaluated the performance of a re-identification system based on different combinations of features using five datasets. Two of these datasets are standard in the re-id literature: iLIDS-VID and PRID 2011. In addition to these, we also used three new challenging datasets that we compiled to better evaluate the re-id performance in "appearance impaired" scenarios.

### 4.1. Standard Datasets

The **iLIDS-VID** dataset [33] is a random collection of 300 persons from the iLIDS Multiple-Camera Tracking Scenario [1]. For each person there are two cropped (64 × 128 pixels/frame) image sequences, captured from two non-overlapping cameras. The lengths of the sequences vary from 23 to 192 frames. Since the data was collected at an airport with crowded environments, the targets are often partially occluded.

The **PRID 2011** dataset [13] consists of cropped (64 × 128 pixels/frame) sequences of 385 persons from camera A and 749 persons from camera B. For the sake of a fair comparison, as proposed by [33], we only use sequences of 178 persons that have more than 21 frames. The scenes in this dataset are less crowded but the illumination variance is fairly large.

### 4.2. Appearance-Impaired Datasets

To illustrate the need for dynamic-based features we collected three more challenging "appearance impaired" datasets. Two of them consist of video sequences of people wearing black/dark clothing. They are subsets of the iLIDS-VID and PRID 2011 datasets and we named them **iLIDS-VID BK** and **PRID 2011 BK**, respectively. The third dataset, named the **Train Station dataset (TSD)**, has sequences of persons with different clothing and accessories.

The BK extension datasets were collected from the orig-inal datasets by manually selecting persons wearing black clothing. We collected 97 and 35 identities for iLIDSVID BK and PRID BK, respectively.

The **Train Station dataset** (TSD), was collected with a single HD surveillance camera mounted at a public train station in the US. Figure 1 shows sample frames from this set. The dataset has 81 sequences, including 9 targets with 3 sequences wearing different clothing and 54 sequences of randomly selected distractors. The length of the sequences vary from 41 to 451 frames. Cropped images of pedestrians are normalized to $64 \times 128$. While all the sequences were captured by the same camera, the relative viewpoint varies significantly when persons enter, re-enter and exit the scene.

## 5. Experiment protocol

To evaluate the merits of the proposed dynamic based features, we do not apply any supervised metric learning method in the experiments. All ranking results are obtained directly by using the Euclidean distance between feature vectors. For a fair comparison, we only use the training data to learn the GMM model for the dynamic features (DynFV) and the local descriptor (LDFV) features (see Sec. 5.1 for more details). For the iLIDSVID and PRID datasets, we follow the protocol in [33]. In this protocol, each dataset is randomly sampled into two equal size subsets. The sequences from the first camera are used as the probe set while sequences from the second camera are used as the gallery set. We repeat the experiment 10 times and report the average value[2]. For the BK extension datasets, because of the sizes of the datasets are fairly small, we randomly pick the same size of training data in the non-BK part (i.e. 89 and 150 persons for PRID and iLIDSVID respectively) and run two experiments by fixing a different camera as the probe set. For the TSD dataset, we only use the distractors to learn the GMM model. During the testing, we randomly pick one sequence for each target combining with all distractors to form the gallery set. This procedure is repeated 10 times. All experiments for DynFV and LDFV are repeated 10 times to remove the uncertainty in the GMM learning step. Please note that we do not need any ground truth in the feature extraction step.

### 5.1. Features

We compare unsupervised re-id performance when using different combinations of features. We used five different types of features, which are described below. Three of them are *purely spatial-based* features: Local Descriptors encoded by Fisher Vector (LDFV) [25], Color & LBP [14] and Hist & LBP [35]; one is a *mixed local spatio-temporal-based* feature: histogram of Gradients 3D (HOG3D) [15]; and one is our proposed *purely dynamics-based feature*:

[2]We directly use the partition file from the project page of [33].

Dynamics-based features Fisher Vector encoded (DynFV). In all cases, before extracting the features, every frame was normalized to $64 \times 128$ pixels. To remove the impact from background, a mask was applied for each frame. The mask was learned for each camera separately, as follows. First, semantic edges were obtained for each frame from the training sequences using a structured forest [8]. Then, the resulting images were averaged followed by a thresholding step to obtain a region covering semantic edges with high average scores.

#### 5.1.1 Spatial-based Features

**LDFV**, proposed in [25], has shown competitive performance in the single shot re-id problem. For each pixel, a 7-dimension $D$ local descriptor is computed for each color channel of HSV: $D = \{x, y, I, I_x, I_y, I_{xx}, I_{yy}\}$ where $x, y$ are the coordinates of the pixel, $I$ represents the intensity value and $I_x, I_{xx}$ are the first and second order gradient on the $x$ direction. Different from the original work, we only use one time $x$ and $y$ since they are redundant for the different color channels. Finally, the resulting 17-dimensional local feature vectors are encoded using Fisher vectors based on a GMM with 16 Gaussians. Following the suggestion in [35], we extract LDFV from 75 $21 \times 16$ patches with $50\%$ overlap.

**Color&LBP** [14] features are extracted from 217 $16 \times 8$ patches with $50\%$ overlap where each color channel from $\{H, S, V, L, A, B\}$ is represented by its mean value and local texture is represented by LBP codes[27].

**Hist&LBP** was suggested by [35]. We ran experiments using the same spatial splitting with LDFV and adopt histograms with 16 bins to represent the 8 color channels $\{R, G, B, H, S, V, Y, U\}$ and apply LBP codes to represent local textures. All spatial features are extracted on each frame independently, followed by an average pooling in time to obtain the final feature vector.

To evaluate the effectiveness of the different types of features, we further split the spatial-based features into a color portion and a texture portion, which include color histogram (Hist) and LBP from Hist&LBP, color mean within small patch (Mean) from Color&LBP and the Fisher vector encoded color (LDFV-color) and edge (LDFV-edge) from LDFV.

#### 5.1.2 Motion-based Features

**HOG3D** [15] is a local feature that captures *both* spatial and temporal gradients. To extract these features, we follow [33], where each video sequence is split into $42 \times 42 \times 10$ even cells with $50\%$ overlap. Then, average pooling is applied along the time dimension. The final feature vector is formed by concatenating these averaged vectors from each cell.

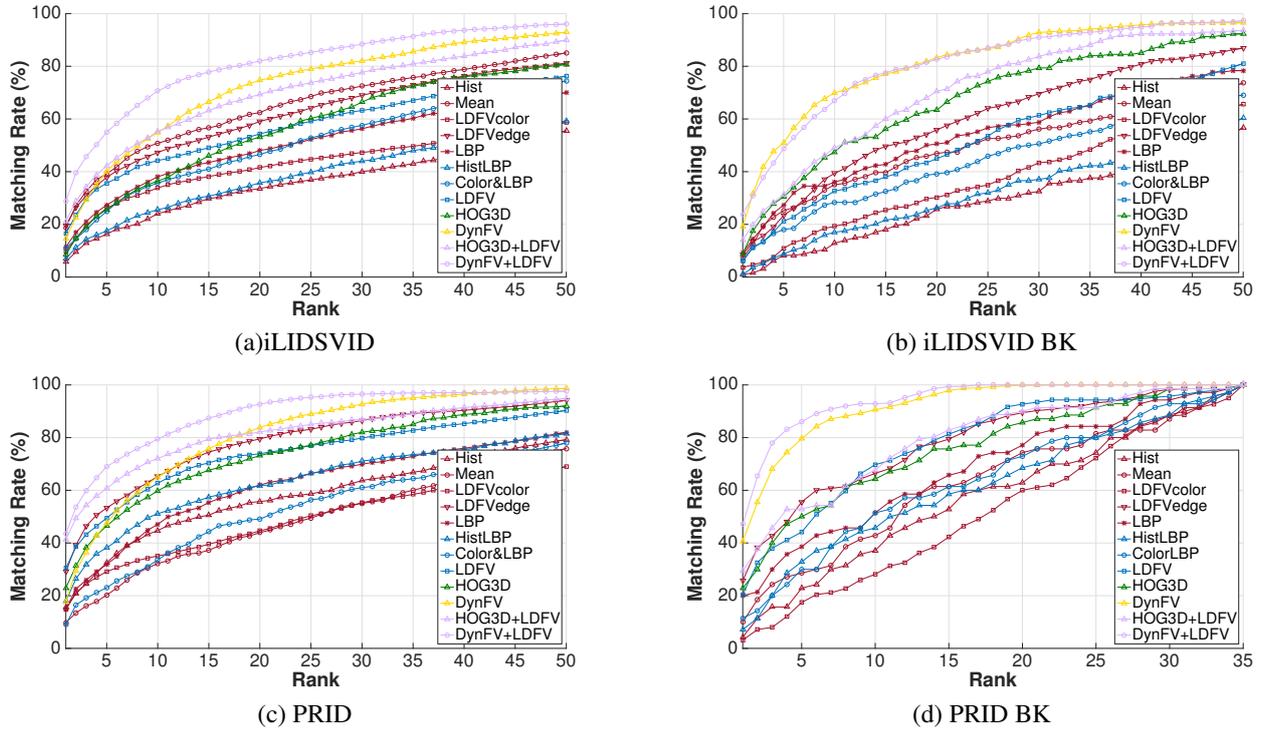

Figure 5. CMC curves for iLIDSVID, PRID and the BK extension datasets

**DynFV** are the proposed dynamics-based Fisher vector encoded features: for each target, we extract sets of dense trajectories (15 frames long) using the code provided by [32], in 18, $32 \times 36$ grids with $50\%$ overlap. We use sliding windows of length $a_i \in \{5, 9, 14\}$. The number of Gaussians for Fisher vector encoding is set to 12.

.

### 5.1.3 Feature Fusion

We apply a *simple* score-level fusion when combining different features. Before fusion, a min-max normalization is used to map the distance range to $[0, 1]$.

$$d_f^{norm} = \frac{d_f - min(d_f)}{max(d_f) - min(d_f)} \qquad (4)$$

where $d_f$ is the Euclidean distance for feature $f$. Then, the final distance will be the summation of $d_f^{norm}$.

## 6. Experiments and Results

Next we present a series of experiments and discuss the results. In all cases we measure performance by comparing Ranking scores. Furthermore, for the analysis of the merits of the features we also give CMC plots and report the proportion of uncertainty removed (PUR) [29] score:

$$PUR = \frac{\log N + \sum_{r=1}^{N} M(r) \log M(r)}{\log N}$$

where $M(r)$ is the value of the match characteristic at rank $r$ and $N$ is the size of the gallery test.

### 6.1. Feature Analysis

In these experiments we studied how using different types of features affects re-id performance. The results are reported in tables 1 to 3, where each row shows the performance when using a different (sub)set of features. Features are grouped as follows. Rows 1-5: a single component (i.e. color *or* texture) of a spatial-based feature; rows 6-8: spatial-based features; rows 9-10: features that incorporate temporal information; and rows 11-12: combinations of spatial-based and temporal features. In all cases, the best performance in each group is shown bold.

**iLIDSVID (BK), PRID (BK) Sets:**

Figure 5 shows the CMC curves and Tables 1, 2 show the re-id performance scores for the iLIDSVID and iLIDSVID BK, and for the PRID and PRID BK datasets, respectively.

**Spatial-based Features-Components:** Color mean achieves the best performance among the three color features for the iLIDSVID, iLIDSVID BK and PRID BK

Table 1. Results for iLIDSVID dataset and iLIDSVID BK dataset

|  | iLIDSVID | | | | | iLIDSVID BK | | | | |
|---|---|---|---|---|---|---|---|---|---|---|
| Feature | 1 | 5 | 10 | 20 | PUR | 1 | 5 | 10 | 20 | PUR |
| Histogram[35] | 5.87 | 16.33 | 24.07 | 33.60 | 6.50 | 1.03 | 8.25 | 12.89 | 25.77 | 7.30 |
| Mean[14] | 18.73 | 39.20 | 50.80 | 62.40 | 23.78 | 7.22 | 25.26 | 35.05 | 46.91 | 11.37 |
| LDFV-color[25] | 9.71 | 26.19 | 33.81 | 41.59 | 8.85 | 3.61 | 10.93 | 19.28 | 30.21 | 4.19 |
| LDFV-edge[25] | **19.59** | **37.81** | **47.29** | **59.07** | 20.78 | 8.45 | 23.40 | 39.38 | 55.88 | 12.59 |
| LBP[27] | 11.40 | 27.33 | 38.07 | 48.07 | 12.29 | **9.28** | **27.32** | **36.08** | **50.52** | **15.10** |
| HistLBP[35] | 6.93 | 17.53 | 25.53 | 35.73 | 7.30 | 1.03 | 8.76 | 17.01 | 26.29 | 5.88 |
| ColorLBP[13] | 10.93 | 24.87 | 35.67 | 46.53 | 14.16 | **6.19** | 18.04 | 28.35 | 39.18 | 8.75 |
| LDFV[35] | **16.55** | **35.64** | **44.24** | **54.33** | **17.82** | 6.08 | **21.24** | **32.68** | **44.95** | **10.26** |
| HOG3D[15] | 8.47 | 25.60 | 36.47 | 53.20 | 15.89 | 8.76 | 30.41 | 47.42 | 63.40 | 20.94 |
| **DynFV (Ours)** | 14.65 | 40.14 | 54.89 | 74.71 | 26.90 | 19.18 | 50.88 | 69.90 | 83.25 | 31.35 |
| HOG3D+LDFV | 21.27 | 42.21 | 55.24 | 69.28 | 26.66 | 14.12 | 31.34 | 49.07 | 70.52 | 21.82 |
| DynFV+LDFV | **28.76** | **54.97** | **70.61** | **81.97** | **37.79** | **23.81** | **48.76** | **67.01** | **82.68** | **30.86** |

Table 2. Results for PRID dataset and PRID BK dataset

|  | PRID | | | | | PRID BK | | | | |
|---|---|---|---|---|---|---|---|---|---|---|
| Feature | 1 | 5 | 10 | 20 | PUR | 1 | 5 | 10 | 20 | PUR |
| Histogram[35] | 15.62 | 32.47 | 44.72 | 55.73 | 12.19 | 4.29 | 22.86 | 37.14 | 62.86 | 7.73 |
| Mean[14] | 9.78 | 20.22 | 32.25 | 43.93 | 6.66 | 10.00 | 28.57 | 42.86 | 74.29 | 11.50 |
| LDFV-color[25] | 14.70 | 29.12 | 35.10 | 44.61 | 8.74 | 3.14 | 17.43 | 28.14 | 60.00 | 3.32 |
| LDFV-edge[25] | **29.06** | **53.26** | **65.30** | **79.53** | **29.12** | **25.71** | **55.57** | **66.43** | **89.43** | **20.73** |
| LBP[27] | 14.83 | 31.91 | 47.08 | 62.02 | 13.65 | 20.00 | 38.57 | 51.43 | 77.14 | 16.66 |
| HistLBP[35] | 18.0 | 38.2 | 51.1 | 61.8 | 16.1 | 7.14 | 32.86 | 45.71 | 68.57 | 7.01 |
| ColorLBP[14] | 8.99 | 23.03 | 33.60 | 49.10 | 8.54 | 11.43 | 30.00 | 51.43 | 72.86 | 11.43 |
| LDFV[25] | **30.45** | **49.48** | **62.67** | **73.87** | **26.83** | **20.43** | **44.14** | **69.71** | **92.71** | **18.02** |
| HOG3D[15] | 22.92 | 46.52 | 59.78 | 73.15 | 23.48 | 22.86 | 50.00 | 64.29 | 85.71 | 20.61 |
| **DynFV (Ours)** | 17.63 | 47.54 | 65.00 | 83.85 | 27.61 | 40.57 | 79.57 | 90.57 | 99.86 | 42.07 |
| HOG3D+LDFV | 41.33 | 60.70 | 72.07 | 82.22 | 37.27 | 29.43 | 52.86 | 67.71 | 90.00 | 25.86 |
| DynFV+LDFV | **43.57** | **68.99** | **79.39** | **92.67** | **44.85** | **47.14** | **86.00** | **92.86** | **100.00** | **50.63** |

datasets, but LDFV-edge outperforms color and LBP in the iLIDSVID, PRID and PRID BK datasets.

**Spatial-based Features:** Rows 6-8 show the performance of spatial-based features commonly used in the re-id literature. Out of these features, LDFV gives the best performance. The reason for the superior performance of the LDFV features is two-fold. First, in general, Fisher vector encoding performs better than average pooling and histogram; and second, because the data consists of multiple frames, LDFV has many samples to get better estimates of the underlying GMM. As expected, the performances for all the spatial features, except LBP and ColorLBP in PRID BK dataset, decrease notably when the features are used on the appearance impaired BK datasets, especially considering that these datasets have smaller galleries.

**Motion-based Features:** HOG3D is a spatial *and* temporal based feature. Its performance is significantly lower than when using purely spatial-based features. However, performance does not degrade when used in the BK sets. These results suggest that the temporal component of this feature helps distinguishing different targets with similar appearance. It should be noted that we obtain better accuracy using HOG3D in the PRID dataset than the results reported in Table 4 of [33]. A reason for this is that instead of using only 4 uniformly sampled candidate fragments, we use all HOG3D features from dense sampled cells to do average pooling, which can provide more stable and less noisy features. In the original datasets, DynFV has Rank 1 performance worse than LDFV but similar to HistLBP and HOG3D and better or similar PUR performance among all single type features, which is remarkable since DynFV does not use any type of appearance based information. In the appearance deprived sets, the DynFV performance increases and significantly outperforms all spatial-based and is almost twice as better than HOG3D.

**Feature Fusion:** The last two rows show that the performance improves when using together spatial-based LDFV and temporal-based features. As seen in both tables, joint use of DynFV and LDFV gives much better results in the original iLIDVID and PRID datasets. These results show that DynFV can be used as a powerful complementary feature in video sequences-based re-id. More precisely, using this combination provides a relatively improvement at Rank 1 performances of 73.78% and 43.09% with respect to using using the LDFV feature alone. On the other hand, including spatial-based features in the impaired datasets, also increases performance but the improvement is smaller, with a relative improvement, at the Rank 1 performance, of 24.14% and 16.19%, respectively.

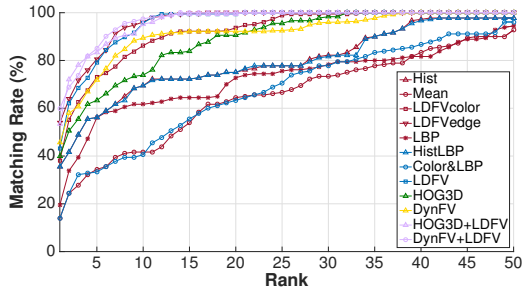

Figure 6. CMC curve for TSD dataset.

Table 3. Results for TSD dataset.

|  | TSD | | | | |
| --- | --- | --- | --- | --- | --- |
| Feature | 1 | 5 | 10 | 20 | PUR |
| Histogram[35] | 35.56 | 56.11 | 69.44 | 75.00 | 35.94 |
| Mean[14] | 13.89 | 34.44 | 41.67 | 64.44 | 17.38 |
| LDFV-color[25] | 38.11 | 73.00 | 86.22 | 93.67 | 45.53 |
| LDFV-edge[25] | **53.89** | **80.56** | **96.11** | **100.00** | **58.19** |
| LBP[27] | 19.44 | 56.11 | 61.67 | 72.78 | 27.54 |
| HistLBP[35] | 35.56 | 56.11 | 69.44 | 75.00 | 36.61 |
| ColorLBP[14] | 13.89 | 33.33 | 40.56 | 63.33 | 18.08 |
| LDFV[25] | **43.11** | **79.11** | **95.67** | **99.44** | **54.34** |
| HOG3D[15] | 40.00 | 63.33 | 73.89 | 90.56 | 42.41 |
| **DynFV (Ours)** | **45.61** | 71.89 | 89.44 | 92.22 | **48.46** |
| HOG3D+LDFV | 53.44 | 83.44 | 95.22 | 99.44 | 60.37 |
| DynFV+LDFV | **59.89** | **85.00** | **97.44** | **100.00** | **62.71** |

• **TSD Set:**
Figure 6 and Table 3 show the CMC curves and performance scores for the TSD dataset, respectively. Since several targets in the dataset only partially change their appearance, spatial features perform fairly well in this set. LDFV-edge performs the best among all single type features. Combining LDFV with DynFV features, provides a relative improvement of performance at Rank 1 of 38.92%.

### 6.2. Effect of Length of Sliding Window

This experiment evaluates the impact of the size of the sliding window used to generate the pyramid of dense trajectories for DynFV. Table 4 shows that the performance at Rank 1 accuracy on the PRID BK dataset has a relative increase of 32.71% when using all three pyramid levels $a_i \in \{5, 9, 14\}$ compared to when just using the original trajectories ($a = 14$).

Table 4. Results with different windows on PRID BK dataset

| Rank | 1 | 5 | 10 | 20 | PUR |
| --- | --- | --- | --- | --- | --- |
| $a = 5$ | 36.29 | 74.14 | 92.00 | 100.00 | 40.55 |
| $a = 9$ | 38.71 | 79.86 | 89.29 | 98.57 | 40.82 |
| $a = 14$ | 30.57 | 76.71 | 89.86 | 97.00 | 36.79 |
| $a_i \in \{5, 9, 14\}$ | **40.57** | **79.57** | **90.57** | **99.86** | **42.07** |

### 6.3. Comparison against State-of-art Methods

Finally, we compare the performance of using the best combination of features, i.e. the proposed DynFV and the spatial-based LDFV features together with a simple *nearest-neighbor* (NN) classifier, against the state-of-art methods for the iLIDSVID and PRID datasets as reported in the literature. It should be emphasized that this approach, as opposed to the competing methods, is *unsupervised* since it does not use any labeled data to train the classifier and naively assigns equally weight to the LDFV and DynFV features. Table 5 shows the results of these comparisons. Even without any supervised learning, DynFV+LDFV+NN achieves competitive performance in the iLIDSVID dataset and the best Rank 1 result in the PRID dataset.

## 7. Conclusion

Until now, most re-id state-of-art approaches relied on appearance based features, extracted from a single or a few images. These approaches do not use the videos which are typically available in surveillance applications and are at a disadvantage in appearance impaired scenarios. In this paper, we proposed DFV features to address these limitations and introduced three new challenging appearance impaired re-id datasets. The proposed DFV feature exploits soft-biometrics, encapsulated in short dense trajectories associated with the targets and benefits from the powerful Fisher vector encoding method. Our extensive experiments show that DFV features carry complementary information to previously used features and that combining them with the state-of-art LDFV features, results in a relative performance improvement at Rank1, compared against using LDFV alone, of 74% and 43% for the iLIDSVID and PRID datasets, respectively, and of 24%, 16%, and 11% for the appearance deprived scenarios in the BK and TS datasets, respectively.

Table 5. Results compared against state-of-the-art algorithms

| iLIDSVID | | | | |
| --- | --- | --- | --- | --- |
| Rank | 1 | 5 | 10 | 20 |
| ColorLBP+DVR[33] | 34.5 | 56.7 | 67.5 | 77.5 |
| Salience+DVR[33] | 30.9 | 54.4 | 65.1 | 77.1 |
| AFDA[18] | **37.5** | **62.7** | **73.0** | 81.8 |
| **DynFV+LDFV+NN** | 28.8 | 55.0 | 70.6 | **82.0** |
| PRID | | | | |
| Rank | 1 | 5 | 10 | 20 |
| ColorLBP+DVR[33] | 37.6 | 63.9 | 75.3 | 89.4 |
| Salience+DVR[33] | 41.7 | 64.5 | 77.5 | 88.8 |
| AFDA[18] | 43.0 | **72.7** | **84.6** | 91.9 |
| **DynFV+LDFV+ NN** | **43.6** | 69.0 | 79.4 | **92.7** |